\documentclass{bmvc2k}
\usepackage{amsmath}
\usepackage{multicolumn}
\usepackage{multirow}

\makeatletter 
  \newcommand\figcaption{\def\@captype{figure}\caption} 
  \newcommand\tabcaption{\def\@captype{table}\caption} 
\makeatother
\bmvcreviewcopy{23}
\title{Action Recognition with Joint Attention on Multi-Level Deep Features}
\addauthor{Jialin Wu}{wujl13@mails.tsinghua.edu.cn}{1}
\addauthor{Gu Wang}{wangg12@mails.tsinghua.edu.cn}{1}
\addauthor{Wukui Yang}{yang-wk15@mails.tsinghua.edu.cn}{1}

\addauthor{Xiangyang Ji}{xyji@mail.tsinghua.edu.cn}{1}

\addinstitution{
 Dept of Automation\\
 Tsinghua University\\
 Beijing, China
}

\runninghead{Wu, Wang, Yang, Ji}{Joint Attention on Multi-Level Deep Features}

\def\etal{\emph{et al}\bmvaOneDot}
\usepackage{setspace}
\everydisplay{
\abovedisplayskip=0.\baselineskip  plus.2ex minus.2ex
\abovedisplayshortskip=-0.13\baselineskip plus.2ex minus.2ex
\belowdisplayskip=-0.\baselineskip plus.2ex minus.2ex
\belowdisplayshortskip=.17\baselineskip plus.2ex minus.2ex
}
\begin{document}

\maketitle

\begin{abstract}
We propose a novel deep supervised neural network for the task of action recognition in videos, which implicitly takes advantage of visual tracking and shares the robustness of both deep Convolutional Neural Network (CNN) and Recurrent Neural Network (RNN). In our method, a multi-branch model is proposed to suppress noise from background jitters. Specifically, we firstly extract multi-level deep features from deep CNNs and feed them into 3d-convolutional network. After that we feed those feature cubes into our novel joint LSTM module to predict labels and to generate attention regularization. We evaluate our model on two challenging datasets: UCF101 and HMDB51. The results show that our model achieves the state-of-art by only using convolutional features.
\end{abstract}

\section{Introduction}
Action recognition and description~\cite{simonyan2014two,donahue2015long,peng2014bag,wang2013action,wang2011action} of videos are fundamental tasks and challenges for computer vision. And they have received significant amount of attention in the research community. With the rapid development of deep Convolution Neural Network (CNN) ~\cite{lin2013network}and Recurrent Neural Network(RNN)~\cite{sharma2015action,mnih2014recurrent,sak2014long,srivastava2015unsupervised} recently, lots of state-of-art methods ~\cite{sharma2015action,simonyan2014two,donahue2015long,simonyan2013deep,wang2013action,wang2011action} adopt them to extract features of different single frames. Nevertheless, when we attempt to extend  image classification tasks to action recognition tasks, the key problem is how to obtain temporal information in videos. Current methods can be mainly categorized into two classes. One is to incorporate tracking using either trajectory-based approaches or attention models, the other is to adopt 3-d convolution.

In this paper, we proposed a novel multi-level and deep supervised method using attention model, which is robust to either intra-class variance or inter-class similarity. We also proposed a soft supervised regularization term to enhance performance in our model. Our main contribution are three folds.
\begin{enumerate}
\item[$\bullet$]   Propose a deep supervised neural network using joint attention model and yields state-of-art results in challenging datasets.
\item[$\bullet$]   Propose a novel attention regularization and enhance the performance of both 3d-convnet module and joint LSTM module in our model.
\item[$\bullet$]   Propose two novel joint LSTM structures that can fuse different features adaptively.
\end{enumerate}
The paper is organized as follows: In Section 2, we introduce current approaches to obtain temporal information in videos. Then, we describe our model in detail in Section 3. After that, we show our experiments and evaluation in Section 4 and 5 respectively. Finally, we present our conclusion and potential future work in Section 6.
\section{Related work}
Video recognition research ~\cite{DBLP:journals/corr/BaMK14,sharma2015action,peng2014action,ji20133d,wang2013action,wang2011action} has been largely driven by the advances in image recognition methods. However, the difference between them is that in video classification tasks, it is of great importance to obtain temporal information among different frames. To deal with that, Current methods often integrate tracking in their algorithm or adopt 3d-convolutional network.
\subsection{Integrating tracking in action recognition}
Lots of state-of-art approaches either extract the trajectories ~\cite{wang2015action,simonyan2013deep,wang2013action,wang2011action} of the videos or calculate the attention ~\cite{DBLP:journals/corr/BaMK14,sharma2015action,mnih2014recurrent}  that classifiers should pay on the videos. All the methods above can be seen as exploiting visual tracking in action recognition tasks, because they aim at finding out the important temporal motion information and decreasing the influences from the background jitters and clutters.
\subsubsection{Trajectory-based model}
Human motion and activities are continuous, the most promising current approaches are based on finding out the trajectories of human motion and combining high level hand-crafted features such as stack fisher vector~\cite{peng2014action} and Bag of Visual Word~\cite{peng2014bag}.Recent improvements of trajectory-based hand-crafted representations include compensation of global (camera) motion~\cite{jain2013better,wang2013action} , and the use of the fisher vector encoding or its deeper variant~\cite{ simonyan2013deep}. Those trajectory-based methods explicitly exploit the tracking results and extract features along the trajectories. However, by doing so, some detailed information outside the trajectories is lost and the performance is highly depended on visual tracking results. Also, those methods assume that the most significant parts of images in video recognition tasks are accord with those in visual tracking tasks.
\subsubsection{Attention-based model}
To prevent from losing information outside the trajectories, attention based approaches  ~\cite{DBLP:journals/corr/BaMK14,sharma2015action,mnih2014recurrent} assign weights for every pixels in origin frames or on feature maps, which can not only discriminate the importance of different parts in frames but also afford less risk of losing subtle information. The most promising attention based methods adopt RNN and can be classified into soft attention~\cite{sharma2015action} and hard attention ~\cite{DBLP:journals/corr/BaMK14,NIPS2014_5542} depended on the way they're trained. Soft attention is a deterministic model and is updated by back propagation. While, hard attention models are usually trained using reinforce learning by setting a policy and reward. In soft attention model, there exist methods cascading RNNs to CNNs, using LSTM to determine the weights should be taken as the attention. In~\cite{sharma2015action}, S. Sharma \etal take a deep inspection on the function of LSTM serving as the attention generator. And Jader berg \etal ~\cite{jaderberg2015spatial}  use spatial transform module to deploy the attention. Yeung \etal ~\cite{yeung2015every} use the temporal attention model to do the dense action labelling and report higher accuracy.
\subsection{LSTM}
LSTM module includes a hidden state, memory state, forget, output, and input gates. During the forward process in each time step, we use the hidden state of the last frame and the current input to calculate i, f, o gates and to further control the input and output of the LSTM.

\begin{eqnarray}
&&{i_{t} = \sigma(W_{xi}x_{t}+W_{hi}h_{t-1}+b_i)}\nonumber\\
&&f_{t} = \sigma(W_{xf}x_{t}+W_{hf}h_{t-1}+b_f)\nonumber\\
&&c_{t-\frac{1}{2}} = tanh(W_{xc}x_{t}+W_{hc}h_{t-1}+b_c)\nonumber\\
&&o_{t} = \sigma(W_{xo}x_{t}+W_{ho}h_{t-1}+b_o)\nonumber\\
&&c_{t} = f_{t}\odot c_{t-1} + i_{t}\odot c_{t-\frac{1}{2}}\nonumber\\
&&h_{t} = o_{t}\odot tanh(c_{t})\nonumber
\end{eqnarray}
where $\sigma$ is element-wise sigmoid function and $\odot$ is element-wise multiplication.

Compared with traditional RNNs, LSTM module can learn and deal with longer and more complex three dimensional video signals because of their additional states and gates. And it is also plausible to stack LSTMs on the top of other LSTM to build a network deep both temporally and spatially as discussed in~\cite{donahue2015long} .

\subsection{3d-convolution}
Another way to obtain temporal information is to use 3d convolutional network ~\cite{tran2014learning,ji20133d,yao2015describing}. 3d convolution regards the video as a volume signal rather than a stack of single frames. Compared with simple 2d convolution, the kernels of 3d convolution have an additional time dimension, which can assign weights to different frames and suppress temporal jitters. D. Tran \etal ~\cite{tran2014learning} research on 3d convolution network and find out that gradually pooling space and time information and building deeper networks achieves state-of-art result. And S. Ji \etal~\cite{ji20133d} use the 3d convolution in human action recognition tasks and yield good results.
\section{Model}
\begin{figure}[h]
\centering
   \includegraphics[width=9cm]{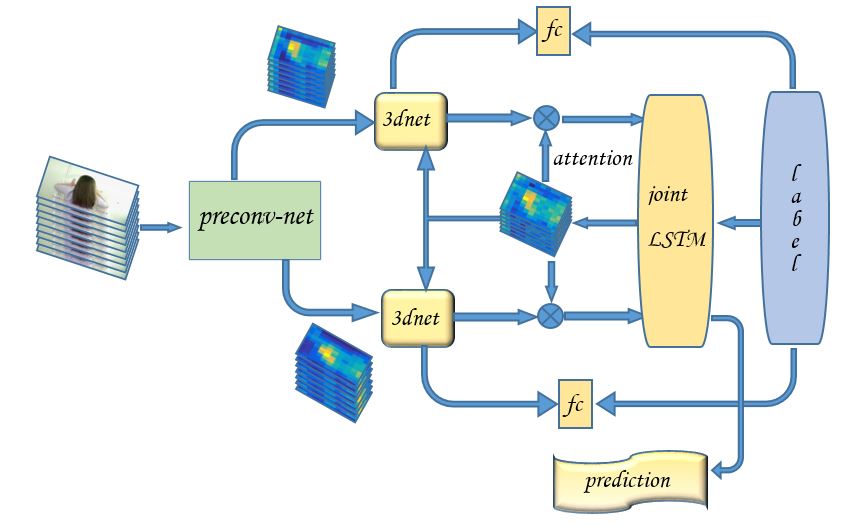}
\caption{Main model including pre-conv net, 3d-convnet and joint LSTM module.}
\end{figure}
This work proposes a multi-branch attention model which adaptively makes use of features from different levels of generality. Our proposed model consists of 3 modules as shown in Figure 1. Firstly, we extract features from different layers in preconv-net, which represent different levels of generality, and then feed those features to 3d-convnet to obtain short-term temporal information. At last, we use the joint LSTM module to obtain long-term temporal information.
\subsection{Pre-conv net}
Considering the strong abilities and robustness of deep Convolutional Neural Network, we adopt VGG 16 network as our pre-conv net to extract features of each frames. The benefits of adopting multi-layer features in visual tracking tasks are partly demonstrated in ~\cite{wang2015visual}, and we observed that in action recognition tasks, different levels of description is also useful for the following reasons. Firstly, there exist a trade-off between excessive and insufficient generality. Through deeper layer we can generally classify the action at the cost of losing lots of specific details. Especially in cases when inter class objects are similar, the classifiers become less robust. Secondly, because of different sizes and resolution of target objects, only a single layer's features will not be sufficient to represent all the targets. Thus, we firstly obtain features from different layers and then fuse them later to avoid losing the most informative and suitable description. 
\subsection{3d-convnet}
Inspired by D. Tran \etal~\cite{tran2014learning}, we use the 3d-convnet to obtain short-term temporal information. However, because we have a joint LSTM module on the top of 3d-convnet module, it is not efficient to train the parameters using gradients back from the LSTM module. Thus we propose a deep supervised strategy to train the 3d-convnet partly independently with an attention regularization term. Specifically, the output of the 3d-convnet are further fed into 2 FC (fully connected) layers to calculate the probabilities for each action label using softmax and updated with the cross entropy loss function. And we observe that the attention we calculate frequently leads the classifier to focus on where feature maps have relatively higher and stronger response as shown in figure 2. Thus we add an attention regularization to encourage the outputs of 3d-convnet to be similar to the attention. The final loss function for the 3d-convnet is defined as below. 
\\\[ \ell = -\sum_{t}ylog(\hat y)+\lambda\sum_{t} \sum^{K^2}_{i=1}(\frac{\sum^{N}_{k=1}x_{t,i,k}}{\sum^{K^2}_{j=1}\sum^{N}_{k=1}x_{t,j,k}}-att_{t,i})^2 +\zeta\sum_{i,j}\theta^2\]
\\where $K^2$ is the spatial size of feature maps and $x_{t,i,k}$ indicates the $k$ th channels of input feature maps in location $i$, time step $t$ , N is dimension of the feature maps  and $\hat y = \frac{exp(W^\top_{3d,i}fc_{2})}{\sum_{j} exp(W^\top_{3d,j}fc_{2})}$ is the softmax probability.
\begin{figure}[h]
\centering
   \includegraphics[width=10cm,height = 3cm]{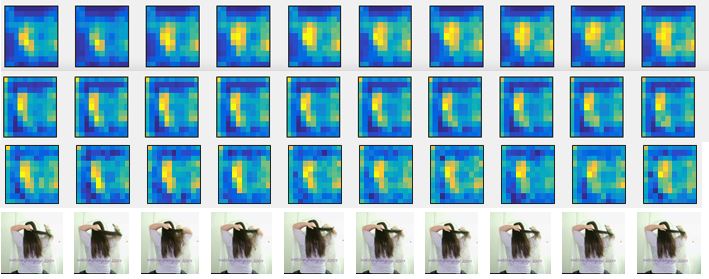}
\caption{The first 2 rows are the feature maps from $conv_{5\_3}$ and $pool_{4}$ layer, the third row is the attention we obtain and the last row is the corresponding raw RGB frames.}
\end{figure}
\subsection{Soft-attention joint LSTM module}
To adaptively fuse multi-level features, we extend origin LSTM model to accept multi-inputs. Specifically, we define different input, forget, and output sub-gates for each input with activate function tanh, which maps the outputs of sub-gates to range (-1, 1). Then the final gates of our joint LSTM are defined as the linear function of corresponding sub-gates with activate function sigmoid, which maps the results to range (0, 1). In our model, we define input, forget, and output gates for the joint LSTM as below, respectively.
\[i_{tk} = tanh(W_{xik}x_{tk}+W_{hik}h_{t-1}+b_{ik})\]
\[f_{tk} = tanh(W_{xfk}x_{tk}+W_{hfk}h_{t-1}+b_{fk})\]
\[o_{tk} = tanh(W_{xok}x_{tk}+W_{hok}h_{t-1}+b_{ok})\]
and
 \[ i_{t} = \sigma(\sum_{k}(W_{iik}i_{tk})) ,\quad f_{t} = \sigma(\sum_{k}(W_{ffk}i_{tk})),\quad o_{t} = \sigma(\sum_{k}(W_{ook}i_{tk}))\]
where $k$ indicates the $k$ th input to our joint LSTM module.
\\Moreover, to model the inputs to the joint LSTM module, we work out 2 different architectures as shown below.
\begin{figure}[h]
\centering
   \includegraphics[width=5cm]{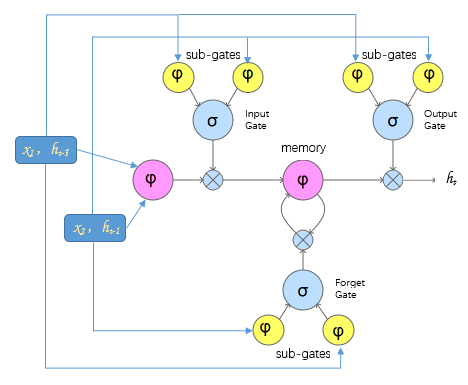}
   \includegraphics[width=5cm]{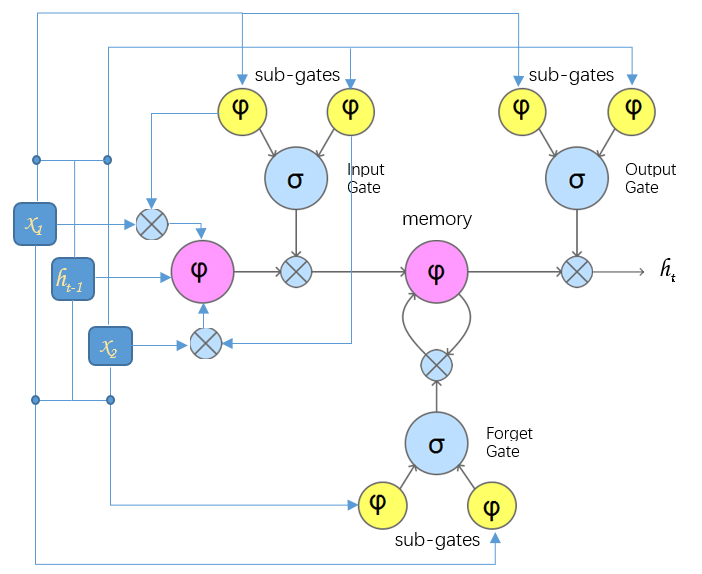}
\caption{Early fusion and late fusion model.}
\end{figure}
\\The first structure, called early fusion, assigns weights to the different inputs by matrices $W_{cxk}$, while, the second structure is more complicated. In structure 2, called late fusion, each input is firstly controlled by their corresponding sub-input gates and then further controlled by the final input gates as they enter the memory state of our proposed joint LSTM module. Given that the $W_{xck}$ are implicitly updated by different inputs, we hope, in our early fusion model, inputs can be successfully weighted by them. And in structure 2, inputs are explicitly controlled by their corresponding sub-input gates. We define the early fusion and late fusion as below respectively.
\begin{eqnarray}
&&early \quad fusion: c_{t-\frac{1}{2}} = tanh((\sum_{k}W_{xck}x_{tk})+W_{hck}h_{t-1}+b_{c})\qquad\qquad\qquad \qquad \qquad  \qquad  \nonumber\\
&&late \quad fusion:c_{t-\frac{1}{2}} = tanh((\sum_{k}W_{xck}(x_{tk}\odot i_{tk}) )+W_{hck}h_{t-1}+b_{c})\qquad \qquad \quad\nonumber 
\end{eqnarray}
Also, it is plausible to have an input from optical flow images. In DoS~\cite{DBLP:journals/corr/VeeriahZQ15} method, the author enhances the model to be more aware of salient dynamic patterns by using the differentiation of the cell of LSTM. And we observe that salient dynamic patterns frequently happen together with strong response in optical flow images. Thus, by adding an inputs from the optical images, our model shall be more aware of higher salient patterns as well. When calculating the attention, we assume the results can be predicted by the hidden state of our LSTM module, and adopt a softmax over the feature maps as shown below. The input is weighted averaged of feature maps.
\[att_{t,i}=p(L_{t}=i|h_{t-1})=\frac{exp(W^\top _{att,i}h_{t-1})}{\sum_{j=1}^{K^2}exp(W^\top_{att,j}h_{t-1})} \quad and\quad  x_{t} = \sum^{K^2}_{i=1}att_{t,i}*X_{t,i}\]
where, $X_{t,i}$ is the feature vector in time step $t$ and location $i$, quad $K^2$ is the spatial size of feature maps, $x_{t}$ is the input feature vector to LSTM module.
\\And we adopt softmax method for prediction and cross entropy loss for training
\[prob(label = i) =  \frac{exp(W^\top_{prob,i}h_{t})}{\sum_{j} exp(W^\top_{prob,j}h_{t})}\quad and\quad \ell = -\sum_{t}ylog(\hat y) +\zeta\sum_{i,j}\theta^2\]
\subsection{Spatial pyramid}
Given that the output feature maps of 3d-convnet are 2-dimensional and contain certain spatial information, if we simply calculate the weighted average of them, the spatial information will be lost. Thus we adopt a spatial pyramid trick as shown in Figure 4.
\begin{figure}[h]
\centering
   \includegraphics[height=4.5cm]{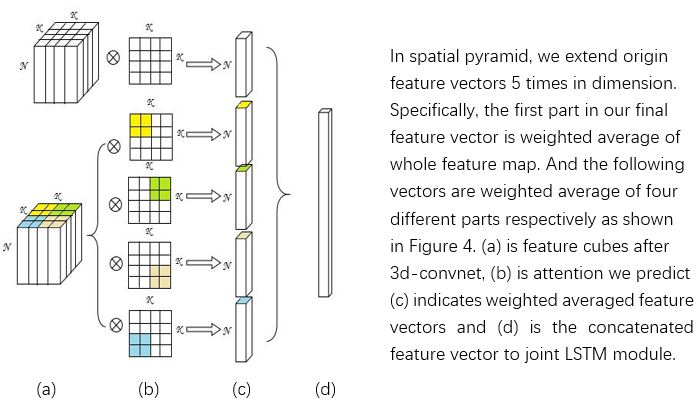}
\caption{Illustration of spatial pyramid.}
\end{figure}
\\Now, the input to the LSTM module is $ x_{t}=[x_{full,t},x_{region1,t},x_{region2,t},x_{region3,t},x_{region4,t}]$
\subsection{Prediction}
To predict the label of an income video we separate the video into several small clips that contain a fixed number of frames and pass them in a random order in train phase. Specifically, we choose the first frame of each different clips in a video from the first frame of the video with a stride of s. For each clip we extract consecutive $L_1$ frames and use the last $L_2$ frames in that clip to predict labels.
\[label = argmax_{i}(\sum^{L_1}_{t=L_1-L_2}\frac{pred(t)}{L_2} ) \]where $ pred(t) = argmax_{i} (prob_{t}(label==i))$
\section{Experiments}
We use a 6G memory 980Ti GPU in our experiments. And due to the memory limitation, we only adopt 2 branches convnet with a batchsize of 64.
\subsection{Datasets}
In experiments, we use HMDB51 and UCF101 datasets. 
\\\textbf{HMDB51}~\cite{kuehne2011hmdb}: The dataset contains 6849 clips divided into 51 action categories and split into 3 parts. Each split of the dataset contains 70 videos for training and 30 for testing.
\\\textbf{UCF101}~\cite{journals/corr/abs-1212-0402}: UCF101 is an action recognition data set of realistic action videos, having 101 action categories. The dataset contains 13K videos from 101 action categories, and is split into 3 different parts with each part containing 9,500 training videos.
\subsection{Configuration}
\textbf{Configuration of pre-conv net}: We adopt the VGG16 network introduced in ~\cite{lin2013network} using caffe~\cite{jia2014caffe}, as its deep convolutional structure is suitable for extracting features representing different levels of generality. And we adopt the $pool_{4}$ layer and $conv_{5\_3}$ layer with the model pre-trained on sport1M dataset then finetuned in each dataset respectively. The raw frames are resized to 384$\times$384 and the feature maps size is 11$\times$11 with 256 channels.
\\\textbf{Configuration of 3d-convnet}:
We choose a two-layer 3d-convnet to obtain short-term temporal information. Each layer has 256 channels in and out with kernel size 3$\times$3$\times$3 in width, height, time respectively, according to ~\cite{tran2014learning}. The $\lambda$, attention term, is set to 1000 and $\zeta$,weights decay penalty, is set to  $10^{-5}$.
\\\textbf{Configuration of LSTM module}:
We adopt a three-layer LSTM to obtain long-term temporal information. The dimension of the memory, hidden, input and output units are set to 1024 for both HMDB51 and UCF101 datasets. And the weight decay coefficient $\zeta$ is set to $10^{-5}$. In prediction, we choose the length of each video clip as $L_1$= 30 frames with a batch size of 64 due to the limitation of our GPU memory, and $L_2$ = 16 to predict labels.
\section{Evaluation}
Firstly, we report our average accuracies on HMDB51 and UCF101 datasets and compare with the state-of-art methods in Table 1.   
\begin{table}[h]
\centering  
\begin{tabular}{|l|c|c|}  
\hline
model & UCF101 &HMDB51\\\hline
Two-stream Conv net~\cite{simonyan2014two}	&88.0\%	&61.4\%
\\\hline 
Soft attention~\cite{sharma2015action}	&-	&41.3\%
\\\hline 
Multi-skip Feature Stacking~\cite{conf/cvpr/LanLLHR15} 	&-	&65.1\%  
\\\hline 
LRCN~\cite{donahue2015long}	&82.9\%	&-
\\\hline 
iDT+FV~\cite{peng2014action}	&85.9\%	&57.2\%
\\\hline 
TDD \cite{wang2015action}&90.3\% &63.2\%
\\\hline
TDD+iDT~\cite{wang2015action}	&91.5\%	&65.9\%

\\\hline 
\textbf{Early fusion joint LSTM(our model)}	&90.3\%	&61.2\%
\\\hline 
\textbf{Late fusion joint LSTM(our model)}	&90.6\%	&61.7\%
\\\hline 
\end{tabular}
\caption{Comparison with the state-of-art models.}
\end{table}

The results show that our best model outperforms original soft attention model about 20.4\% and is better than iDT+FV model and Two-stream model in both datasets. Also, our model is better than TDD in UCF101 and rivals TDD+iDT. However, TDD+iDT model encodes both RGB and flow images' features using Fisher vector, which is more complex and leads to huge computational cost, while our model simply takes RGB images as input.

Secondly, we look deeper into features from VGG16 net as input to our multi-branch network and demonstrate the benefits of adopting different levels of generalities. In action recognition tasks, the trade-off between general and specific is hard to balance. So it is not likely we can use only one single layer's features to obtain sufficient and plausible description for all the videos. By adopting multi-description, our model outperforms single description significantly as shown in Table 2. In the  experiments, when testing single input, we use the model from Sharma \etal ~\cite{sharma2015action}and add the same 3d-convnet module.
\begin{table}[htb] 
  \begin{minipage}[h]{0.4\textwidth} 
   \begin{tabular}{|l|c|}  
\hline
layer  &HMDB51\\\hline
$pool_{4}$	& 42.8\%\\\hline 
$conv_{5\_3}$	&45.5\%\\\hline 
Late fusion 	&61.7\%\\\hline 
\end{tabular}
\caption{Comparison of whether  \protect\\ adopting multi-branch strategy.}
  \end{minipage}%
  \begin{minipage}[h]{0.5\textwidth} 
    \centering
  
\begin{tabular}{|l|c|}  
\hline
straightly connect VGG16 to LSTM	&57.3\%\\\hline 
2d-convnet with regularization &	57.5\%\\\hline 
simple 3d-convnet	&59.4\%\\\hline 
3d-convnet with regularization	&61.9\%\\\hline 
\end{tabular}
\caption{Comparison of whether adopting 3d-convnet.}

  \end{minipage} 
\end{table}

Then, we pay attention to the benefits of adopting 3d-convnet. We compare  3d-convnet with a simple 2d-convnet and straightly feeding feature maps to LSTM module in HMDB51 split1. As the Table 3 shows, 3d-convnet can enhance temporal information for our multi-level model by outperforming 2d-convnet and straightly connecting VGG to LSTM 4.4\% and 4.6\% respectively. In Figure 5, we plot the loss of joint LSTM module. And we observe that with attention regularization, our model converge much more quickly, frequently in the third epoch as shown in Figure 7, to a lower platform.

\begin{figure}[htb] 
  \begin{minipage}[h]{0.43\textwidth} 
    \centering 
    \includegraphics[width=4.5cm]{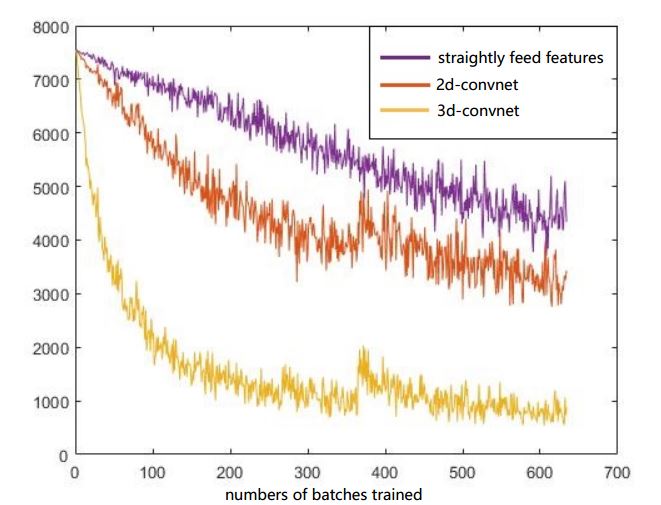}
    \caption{Loss of LSTM.} 
  \end{minipage}%
  \begin{minipage}[h]{0.55\textwidth} 
    \centering
    
\begin{tabular}{|l|c|c|}  
\hline
&early fusion&	late fusion\\\hline
UCF101 split1&90.2\%&	90.7\%\\\hline
UCF101 split2&	90.8\%&	90.8\%\\\hline
UCF101 split3&	89.9\%&	90.3\%\\\hline
UCF101 average&	90.3\%&	90.6\%\\\hline
HMDB51 split1&	60.9\%&	61.9\%\\\hline
HMDB51 split2&	61.4\%&	61.6\%\\\hline
HMDB51 split3&	61.2\%&	61.6\%\\\hline
HMDB51 average&	61.2\%&	61.7\%\\\hline
\end{tabular}
\tabcaption{Comparison of different structures.} 
  \end{minipage} 
\end{figure}

After that, we evaluate 2 different input architectures of our joint LSTM module. In Table 4, we can observe that late fusion model can further improve the performance of our Joint LSTM module. By looking inside, it may because late fusion version discerns the 2 branches of inputs from below more completely and explicitly with early non-linear sub-input gates. Thus, it can fuse different inputs more independently, which leads to higher robustness.

Finally, we provide some examples of the attention we get through videos.
\begin{figure}[h]
\centering
\includegraphics[width=11.5cm]{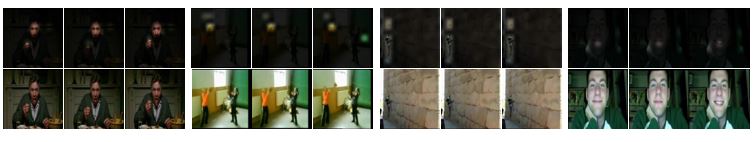}
\\eat \qquad \qquad \qquad catch\qquad\qquad \qquad jump\qquad\qquad\qquad smile\\
\includegraphics[width=11.5cm]{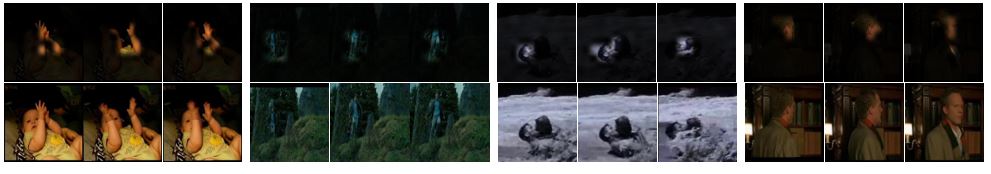}
\\clap \qquad \qquad \qquad run\qquad\qquad \quad \qquad kiss\qquad\qquad\qquad turn\\
\caption{Examples of our attention model. The first row is frames with attenton.Brighter parts receive more and darker receive less. The second row is the corresponding raw frames.}
\end{figure} 

As Figure 6 shows, our joint atention model successfully tracks the most important parts in the vedios, thus classifies them correctly and performs well in most challenging datasets. In the first two rows our model pays more attention on the mouth, the ball, the jumping man and the smile, and in the second two rows, the hands, the legs, the kiss, and the face receive more attetion. Obviously, those parts play important roles when classifing the action into the corresponding classes. 
Figure 7 shows the confusion matrix of our model.
\begin{figure}[h]
\centering
   \includegraphics[width=5cm]{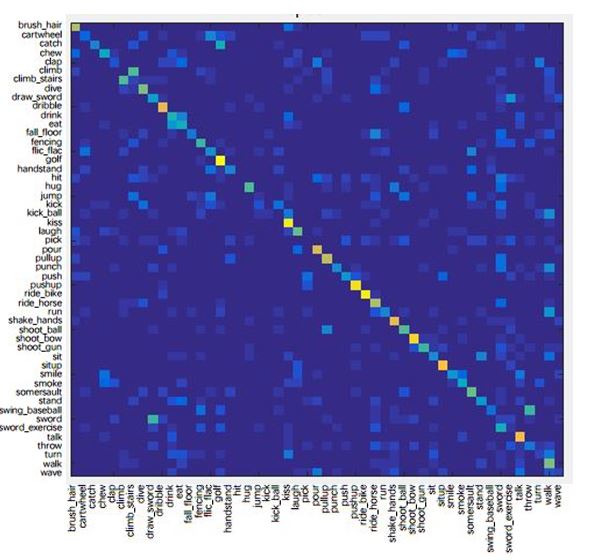}
\qquad
   \includegraphics[width=5cm]{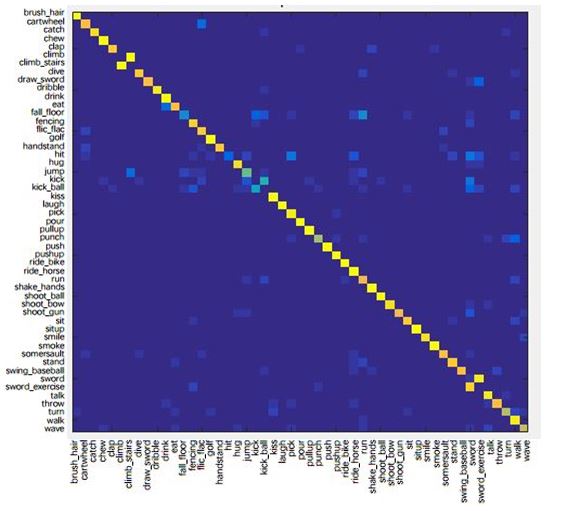}
\caption{Confusion matirx of epoch1 and epoch3 in HMDB51 dataset.} 
\end{figure}

\section{Conclusion and future work}
We have present our model which implicitly takes advantage of visual tracking. Our results on HMDB51 and UCF101 datasets prove that by adopting multi-level of deep features, the performance has been enhanced. What's more, our model can be extended to fuse even more layers and combine more kinds of features to further improve the robustness. And our model can rival the state-of-art by only using convolutional features. Our potential future work is to further extend our model to more tasks in computer vision field like image and video description, etc. Also we are trying to integrate more kinds of features such as Fisher Vector~\cite{peng2014action} and BoVW~\cite{peng2014bag} and explain those features with attention model.
\bibliography{bib}

\end{document}